\documentclass{sig-alternate-05-2015}

\usepackage{tikz}
\usepackage{amsmath}
\usetikzlibrary{fit,positioning}
\usepackage[]{algorithm2e}
\usepackage{booktabs}
\usepackage{bm}
\newcommand{\senLDA}{\textit{senLDA }}

\begin{document}


%


%

\title{On a Topic Model for Sentences}

\numberofauthors{1} 
%
\author{
\alignauthor
Georgios Balikas,$^\dagger$\titlenote{Also affiliated with: Coffreo, Clermont Ferrand, France} Massih-Reza Amini$^\dagger$ and Marianne Clausel$^\ddagger$\\
       \affaddr{University of Grenoble Alpes}\\
       \affaddr{$^\dagger$Computer Science Laboratory (LIG), $^\ddagger$Applied Mathematics Laboratory (LJK)} \\
       \affaddr{Grenoble, France}\\
       \email{\{FirstName.LastName\}@imag.fr}
}
\date{5 May 2016}
\maketitle
\begin{abstract}
Probabilistic topic models  are generative models that describe the content of documents by discovering the latent topics underlying them. However, the structure of the textual input, and for instance the grouping of words in coherent text spans such as sentences, contains much information which is generally lost with these models. In this paper, we propose \textit{sentenceLDA}, an extension of LDA whose goal is to overcome this limitation by incorporating the structure of the text in the generative and inference processes. We illustrate the advantages of \textit{sentenceLDA} by comparing it with LDA using both intrinsic (perplexity) and extrinsic (text classification) evaluation tasks on different text collections.
\end{abstract}



\keywords{Text Mining; Topic Modeling; Unsupervised Learning}

\section{Introduction}
Statistical topic models are generative unsupervised models that describe the content of documents in large textual collections. Prior research has investigated the application of topic models such as Latent Dirichlet Allocation (LDA) \cite{blei2003latent} in a variety of domains ranging from image analysis to political science. 
Most of the work on topic models assumes exchangeability between words and treats documents in a bag-of-words fashion. As a result, the words' grouping in coherent text segments, such as sentences or phrases, is lost. 

However, the inner structure of documents is generally useful, when identifying topics. For instance, one would expect that in each sentence, after standard pre-processing steps such as stop-word removal, only a very limited number of latent topics would appear. Thus, we argue that coherent text segments should pose ``constraints'' on the amount of topics that appear inside those segments.  

In this paper, we propose \textit{sentenceLDA} (\textit{senLDA}), whose purpose is to incorporate part of the text structure in the topic model. Motivated by the argument that coherent text spans should be produced by only a handful of topics, we propose to modify the generative process of LDA. Hence, we argue that the latent topics of short text spans should be consistent across the units of those spans. In our approach, such text spans can vary from paragraphs to sentences and phrases depending on the task's purpose. Also, note that in the extreme case where words are the coherent text segments, the standard LDA model becomes a special case of \textit{senLDA}. 

In the remainder of the paper we present the \senLDA  and we derive its collapsed Gibbs sampler in Section 2, we illustrate its advantages by comparing it with  LDA on intrinsic (\textit{in vitro}) and extrinsic (\textit{ex vivo}) evaluation experiments using collections of Wikipedia and PubMed articles in Section 3, and we conclude in Section 4.

\section{The proposed model}
A statistical topic model represents the words in a collection of $D$ documents
as mixtures of $K$ ``topics'', which
are multinomials over a vocabulary of size $V$. In the case of LDA, for each document
$d_i$ a multinomial over topics is sampled from a Dirichlet prior with parameters $\boldsymbol{\alpha}$. 
The probability $p(w|z=k)$ of a term $w$, given the topic $k$, is represented by
$\phi_{k,t}$. We refer to the complete $K\times V$ matrix of word-topic 
probabilities as $\Phi$. 
The multinomial parameters $\phi_k$
are again drawn from a Dirichlet prior parametrized by $\boldsymbol{\beta}$. Each observed term $w$ in the collection is drawn from a multinomial for the topic
represented by a discrete hidden indicator variable $z_i$. For simplicity in the mathematical development and notation, we assume symmetric Dirichlet priors but the extension to the asymmetric case is straightforward. Hence, the values of $\alpha$ and $\beta$ are model hyper-parameters.
 
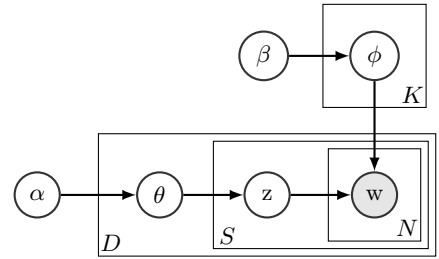
\begin{figure}[t]
\centering
\begin{tikzpicture}
\tikzstyle{main}=[circle, minimum size = 6mm, thick, draw =black!80, node distance = 8mm]
\tikzstyle{connect}=[-latex, thick]
\tikzstyle{box}=[rectangle, draw=black!100]
  \node[main, fill = white!100] (alpha)  {$\alpha$};
  \node[main, node distance = 10mm] (theta) [right=of alpha,] {$\theta$};
  \node[main] (z) [right=of theta,] {z};  
  \node[main, fill = black!10] (w) [right=of z,] {w};
  \node[main, node distance = 12mm] (phi) [above=of w,] {$\phi$ };
  \node[main] (beta) [left=of phi,] {$\beta$ };

  \path (alpha) edge [connect] (theta)
        (theta) edge [connect] (z)
        (z) edge [connect] (w)
        (beta) edge [connect] (phi)
        (phi) edge [connect] (w);
    
  \node[rectangle, inner sep=3mm, fit= (w), draw=black!100, label={[xshift=-4.5mm, yshift=4mm]below right:$N$} ] {};
  \node[rectangle, inner sep=4mm,draw=black!100,fit= (w) (z),label={[xshift=-3mm, yshift=4mm]below left:$S$}] {};
  \node[rectangle, inner sep=5mm,draw=black!100,fit= (theta) (w) (z),label={[xshift=-10mm, yshift=4mm]below left:$D$}] {};
    \node[rectangle, inner sep=3.5mm,draw=black!100,fit= (phi),label={[xshift=-4.5mm, yshift=4mm]below right:$K$}] {};

\end{tikzpicture}
\caption{The \senLDA model. The words $w$ of a sentence share the same topic $z$.}\label{fig:senlda}
\end{figure}

We extend LDA by adding an extra plate denoting the coherent text segments of a document. In the rest, without loss of generality we use sentences as coherent segments. A finer level of granularity can be achieved though, by analysing the structure of sentences and using phrases as such segments. The graphical representation of the \senLDA model is shown in Figure \ref{fig:senlda} and the generative process of a document collection using \senLDA is described in Algorithm \ref{algo:senlda}. 
For inference, we use a collapsed Gibbs sampling method \cite{griffiths2004finding}. 
We now derive the Gibbs sampler equations by estimating the hidden topic variables.

In \senLDA the joint distribution can be factored:
\begin{equation}
p(w,z|\alpha,\beta) = p(w |z,\beta)p(z|\alpha)
\label{eq:factors}
\end{equation}
because the first term is independent of $\alpha$ and the second from $\beta$. 
After standard manipulations as in the paradigm of \cite{heinrich2005parameter} one arrives at: 
\begin{equation}
p(\vec{z}, \vec{w}|\alpha,\beta) = \prod_{z=1}^{K}\frac{\Delta(\vec{n}_z+\beta)}{\Delta(\beta)} \prod_{m=1}^D \frac{\Delta(\vec{n}_m+\alpha)}{\Delta(\alpha)}\label{eq:joint}
\end{equation}
\noindent where $\Delta(\vec{x})=Beta(x_1,\ldots,x_m)=\frac{\prod_{k=1}^{dim \vec{x}} \Gamma(x_k)}{\Gamma(\sum_{k=1}^{dim \vec{x}} x_k)}$ is a multidimensional extension of the beta function used for notation convenience, and $\vec{n}_m$, $\vec{n}_z$ refer to the occurrences of topics with documents and topics with terms respectively. To calculate the full conditional
we take into account the structure of the document $d$ and the fact that  $\vec{w_d}=\{\vec{w}_{d\neg{s}}, \vec{w}_{\neg{s}}\}$, $\vec{z}=\{\vec{z}_{d\neg{s}}, \vec{z}_{\neg{s}}\}$. The subscript $s$ in $\vec{w}_s,\vec{z}_s$ denotes the words and the topic respectively of sentence $s$. For the full conditional  of topic $k$ we have:
\begin{equation}
\begin{split}
p(z_s=k|\vec{z}_{\neg{s}}, \vec{w}) = \frac{p(\vec{w},\vec{z})}{p(\vec{w}, \vec{z}_{\neg{s}})} = \frac{p(\vec{w}|\vec{z})}{p(\vec{w}_{\neg{s}}|\vec{z}_{\neg{s}}) p(w_s)}\frac{p(\vec{z})}{p(\vec{z}_{\neg{s}})}=\\
= \frac{p(\vec{w},\vec{z})}{p(\vec{w}_{\neg{s}}, \vec{z}_{\neg{s}})} \propto
 \frac{\Delta(\vec{n}_z+\beta)}{\Delta(\vec{n}_{z,\neg{s}}+\beta)} \frac{\Delta(\vec{n}_m+\alpha)}{\Delta(\vec{n}_{m,\neg{s}}+\alpha)} 
\end{split}
\label{eq:fullConditional}
\end{equation}

\noindent For the first term of equation Eq. \eqref{eq:fullConditional} we have:
\begin{multline}
  \frac{\Delta(\vec{n}_z+\beta)}{\Delta(\vec{n}_{z,\neg{s}}+\beta)} =
  \frac{ \frac{\prod_{w\in s} \Gamma(\vec{n}_z+\beta)}{\Gamma(\sum_{w\in s}(\vec{n}_z+\beta))}}
  {\frac{\prod_{w\in s} \Gamma(\vec{n}_{z,\neg{s}}+\beta)}{\Gamma(\sum_{w\in s}(\vec{n}_{z,\neg{s}}+\beta))}}= \\ =\prod_{w\in s} (\frac{ \Gamma(\vec{n}_z+\beta)}{\Gamma(\vec{n}_{z,\neg{s}}+\beta)}) \frac{\Gamma(\sum_{w\in s}(n_{z,\neg{s}}+\beta))}{\Gamma(\sum_{w\in s}(n_z+\beta))} = \\
  =   \underbrace{\frac{\overbrace{\prod_{w\in s} (n_{k,\neg{s}}^{(w)}+\beta)\cdots (n_{k,\neg{s}}^{(w)}+\beta+(n_{k,s}^{(w)}-1))  }^{\mbox{A}}}{(\sum_{w\in V} (n_{k,\neg{s}}^{(w)}+\beta))\cdots(\sum_{w\in V} n_{k,\neg{s}}^{(w)}+\beta+(N_{k,s}^{(w)}-1))} }_{\mbox{B}}
\label{eq:inferenceDetails}
\end{multline}
Here, for the generation of A and B  we used the recursive property of the $\Gamma$ function: $\Gamma(x+m)= (x+m-1)(x+m-2)\cdots(x+1)x\Gamma(x)$; $w$ is a term that can occur many times in a sentence and $n_{k,s}^{(w)}$ denotes $w$'s frequency in sentence $s$ given that the sentence $s$ belongs to topic $k$; $N_{k,s}^{(w)}$ denotes how many words of sentence $s$ belong to topic $t$. 

\RestyleAlgo{boxruled}
\begin{algorithm}[t]\small
\DontPrintSemicolon
 \For{ \upshape document $d \in[1,\ldots,D]$}{
  sample mixture of topics $\theta_m \sim$ Dirichlet(a)\;
  sample sentence number $S_d \sim  Poisson (\xi)$  \;
   //Sentence plate\;
  \For{ \upshape sentence $s\in[1,S_d]$}{
  sample number of words  $W_{s} \sim  Poisson (\xi_d)$ \;
  sample topic $z_{d,s} \sim Multinomial(\theta_m)$\;
  //Word plate in each language\;
  \For{ \upshape words $w\in [1, W_{d,s}]$ in sentence $s$ }{
  sample term for $w \sim Multinomial(\phi_{z_{d,s}}) $\;
  }
  } 
 }
\caption{Text collection generation with \textit{senLDA}}\label{algo:senlda}
\end{algorithm}

The development of the second factor in the final step of Eq. \eqref{eq:fullConditional} is similar to the LDA calculations with the difference that the counts of topics per document are calculated given the allocation of sentences to topics and not the allocation of words to topics. This yields:
\begin{multline}
 p(z_s=k|\vec{z}_{\neg{s}}, \vec{w}) = (n_{m,\neg{s}}^{(k)} + \alpha) \times \\ \times \frac{\prod_{w\in s} (n_{k,\neg{s}}^{(w)}+\beta)\cdots (n_{k,\neg{s}}^{(w)}+\beta+(n_{k,s}^{(w)}-1))}{(\sum_{w\in V} (n_{k,\neg{s}}^{(w)}+\beta))\cdots(\sum_{w\in V} n_{k,\neg{s}}^{(w)}+\beta+(N_{k,s}^{(w)}-1))}  
\label{eq:finalMonolingual}
 \end{multline}
where $n_{m,\neg{s}}^{(w)}$ denotes the number of times that topic $k$ has been observed with a sentence from document $d$, excluding the sentence currently sampled. 
Note that Eq. \eqref{eq:finalMonolingual} reduces to the standard  LDA collapsed Gibbs sampling inference equations if the coherent text spans are reduced to words.

The idea of integrating the sentence limits in the LDA model has been previously investigated. For instance, in \cite{wang2009multi} in the context of summarization the authors combine the unigram language model with topic models over sentences so that the latent topics are represented by sentences instead of terms. In \cite{chen2010adaptation} the notion of \textit{sentence topics} is introduced and they are sampled from separate topic distributions and co-exist with the word topics. Also, Boyd et al. \cite{boyd2009syntactic} propose an adaptation of topic models to the text structure obtained by the parsing tree of a document. Our method resembles these works in that it integrates the notion of sentences to extend LDA. In our case though, we directly extend LDA maintaining the association of words to topics, we retain its simplicity without adding extra hyper-parameters thus allowing a fast, gibbs sampling inference, and we do not require any language-dependent tools such as parsers.

\section{Empirical results}

We conduct experiments to verify the applicability and evaluate the performance of \senLDA  compared to LDA. The process is divided into two steps: (i) the training phase, where the topic models are trained to learn the their parameters, and (ii) the inference phase that is for new, unseen documents their topic distributions are estimated. We use the Gibbs sampling inference approach given by Eq. \eqref{eq:finalMonolingual}. The hyper-parameters $\alpha$ and $\beta$ are set to $\frac{1}{K}$, with $K$ being the number of topics. Table \ref{table:trainData} shows the  datasets we used. They come from the publicly available collections of Wikipedia \cite{partalas15lshtc} and PubMed \cite{tsatsaronis2015overview}. The first four datasets (WikiTrain* and PubMedTrain*) were used for learning the topic model parameters; they differ in their respective size. Also, the vocabulary of the PubMed datasets is significantly larger due to the medical terms that appear. During preprocessing we only applied lower-casing, stop-word removal and lemmatization using the WordNet Lemmatizer.\footnote{The code and the data are publicly available  at \url{https://github.com/balikasg/topicModelling/}}  The rest of the document collections of Table \ref{table:trainData} are used for classification purposes and are discussed later in the section. 

\begin{table}[t]
\small\centering
 \begin{tabular}{lcccc}
  \toprule
  &Documents & $|V|$ & Classes & Timing (sec) \\\midrule
  WikiTrain1 & 10,000 & 46,051 & - &182|271 \\
  WikiTrain2 & 30,000 & 65,820 & - &332|434 \\
  PubMedTrain1 &10,000& 55,115 & - &304|433 \\
  PubMedTrain2 & 60,000& 150,440 & - & 1830|2799 \\
  Wiki37& 2,459 & 23,559& 37 & -\\
  Wiki46& 3,657 & 27,914& 46 & -\\
  PubMed25 & 7,248 &40,173 & 25 & -\\
  PubMed50 & 9,035 &47,199 & 50 & -\\
\bottomrule
 \end{tabular}
\caption{Description of the data used after pre-processing. ``Timing'' refers to the 25 first training iterations with the left (resp. right) values corresponding to \senLDA (resp. LDA).}\label{table:trainData}
\end{table}

\noindent\textbf{Intrinsic evaluation} Topic model evaluation has been the subject of intense research. For intrinsic evaluation we report here perplexity \cite{azzopardi2003investigating}, which is probably the dominant measure for topic models evaluation in the bibliography. The perplexity of $d$ held out documents given the model parameters $\vec{\vartheta}$ is defined as the reciprocal geometric mean of the token likelihoods of those data, given the parameters of the model:
\begin{equation}
 p(w_{\text{heldOut}}) = \exp-\frac{\sum_{i=1}^d \sum_{j=1}^{w_i} \log p(w_{i,j}|\vec{\vartheta})}{\sum_{i=1}^d \sum_{j=1}^{w_i} 1}  
\end{equation}
Note that \senLDA  samples per sentence and thus results in less flexibility at the word level where perplexity is calculated. Even though, the comparison between \senLDA and LDA, at word level using perplexity, gives insights in the relative merits of the the proposed model. 
\begin{figure}[t]
\centering
 \includegraphics[width=0.47\textwidth]{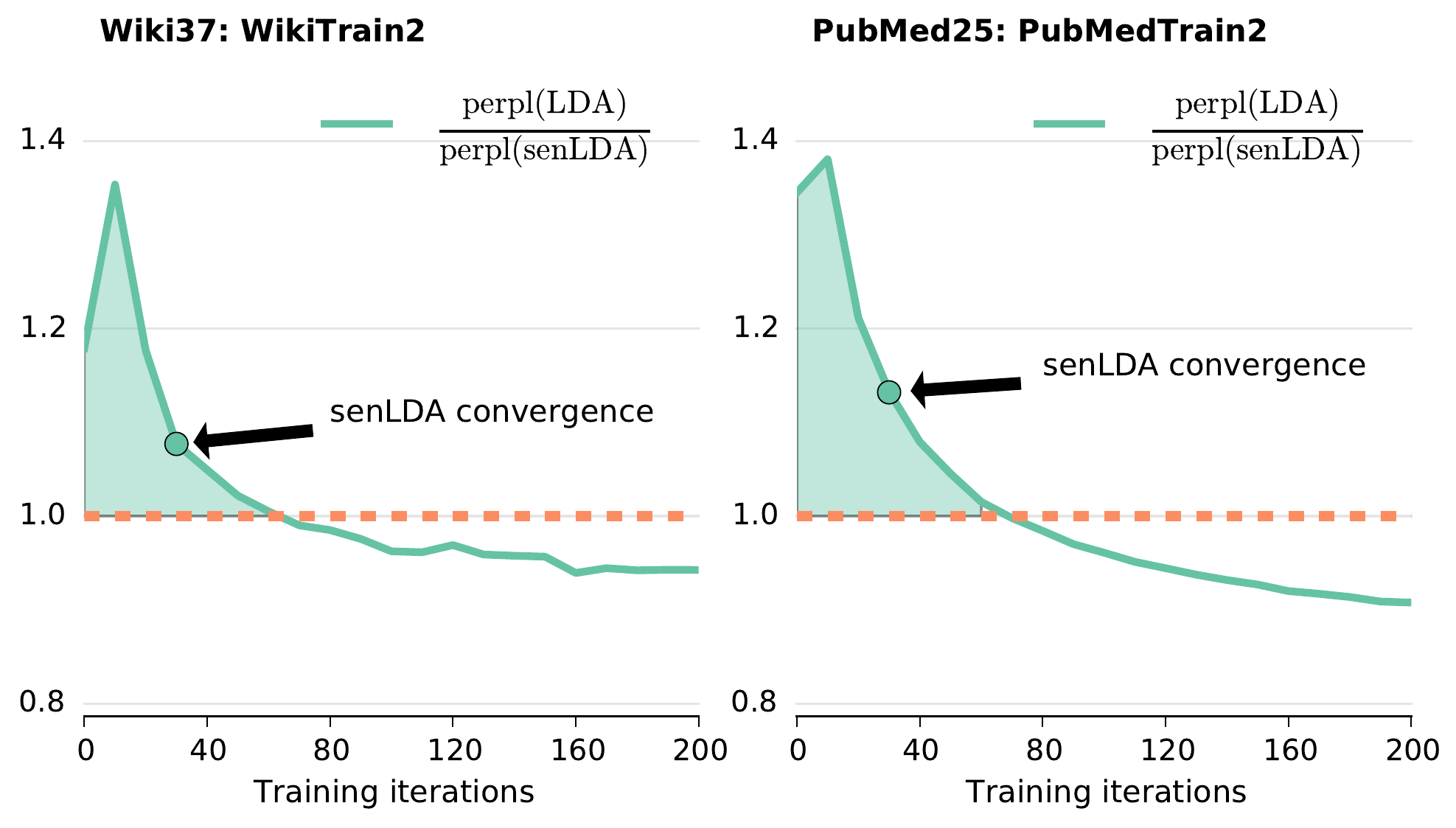}
 \caption{The ratio of perplexities of \senLDA and LDA calculated on Wiki37 and PubMed25.  }\label{fig:perplexity}
\end{figure}

Figure \ref{fig:perplexity} depicts the ratio of the perplexity values between \senLDA and LDA. We set $K=125$ after grid searching $K\in \{25, 75, 125, 175\}$ for perplexity with 5-fold cross-validation on the training data. Values higher (resp. lower) than one signify that \senLDA achieves lower (resp. higher) perplexity than LDA. The figure demonstrates that in the first iterations before convergence of both models, \senLDA performs better. What is more, \senLDA converges after only around 30 iterations, whereas LDA converges after 160 iterations on Wikipedia and 200 iterations on the PubMed  datasets respectively. We define convergence as the situation where the model's perplexity does not any more decrease over training iterations. The shaded area in the figure highlights the period while \senLDA performs better. It is to be noted, that although competitive, \senLDA  does not outperform  LDA given unlimited time resources. However, that was expected since for \senLDA the training instances are sentences, thus the model's flexibility is restricted when evaluated against a word-based 
measure. 

\begin{figure*}[ht]
  \centering
 \includegraphics[width=0.90\textwidth]{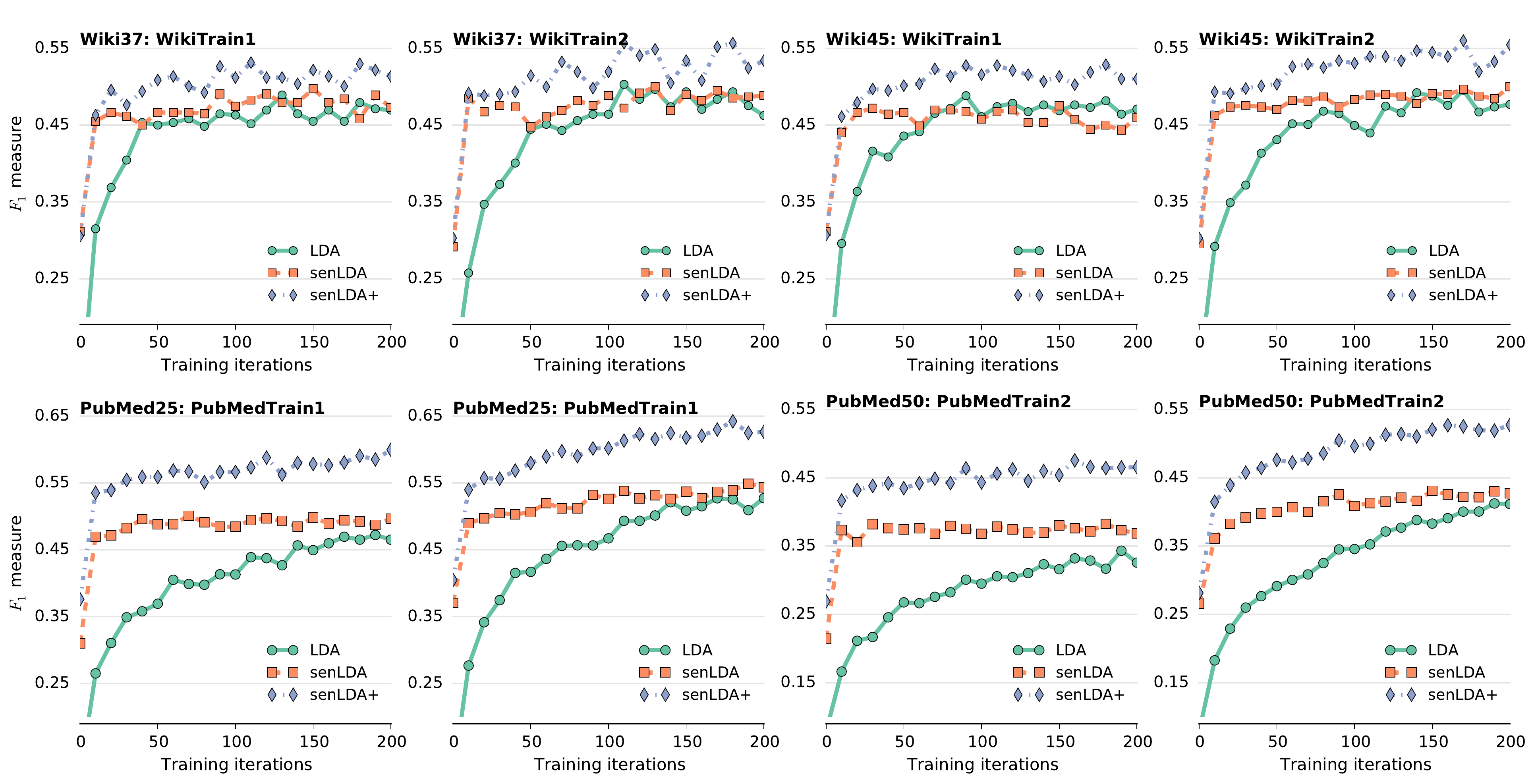}
 \caption{Classification performance on PubMed and Wikipedia text collections using $F_1$ measure.}\label{fig:classfication}
\end{figure*}

An important difference between the models however, lies in the way they converge. From Figure \ref{fig:perplexity} it is clear that \senLDA converges faster. We highlight this by providing exact timings for the first 25 iterations of the models (column ``Timing'' of Table \ref{table:trainData}) on a machine using an Intel Xeon CPU E5-2643 v3 at 3.40GHz. For both models we use our own Python implementations with the same speed optimisations.  Using ``WikiTrain2'' and 125 topics, for 25 iterations the \senLDA needs 332 secs, whereas LDA needs 434 sec., an improvement of 30\%. Furthermore, comparing the convergence, \senLDA needs 332 secs (25 iterations) whereas LDA needs more than 2770 secs (more than 160 iterations) making \senLDA more than 8 times faster. Similarly for the ``PubMedTrain2'' dataset which is more complex due to its larger vocabulary size, \senLDA converges around 12 times (an order of magnitude) faster.
Note that \textit{senLDA}'s fast convergence is a strong advantage and can be highly appreciated in different application scenarios where unlimited time resources are not available.  

\noindent\textbf{Extrinsic evaluation} Previous studies have shown that perplexity does not always agree with human evaluations of topic  models \cite{azzopardi2003investigating} and it is recommended to evaluate topic models on real tasks. To better support  our development for \senLDA applicability we also evaluate it using text classification as the evaluation task. 
 For text classification, each document is represented by its topic distribution, which is the vectorial input to Support Vector Machines (SVMs). The classification collections are split on train/test (75\%/25\%) parts. The SVM regularization hyper-parameter $\lambda$ is selected from $\lambda\in[10^{-4},\ldots,10^4]$ using 5-fold cross-validation on the training part of the classification data.
The PubMed testsets are multilabel, that is each instance is associated with several classes, 1.4 in average in the sets of Table \ref{table:trainData}. For the multilabel problem with the SVMs we used a binary relevance approach.    
To assess the classification performance, we report the $F_1$ evaluation measure, which is the harmonic mean of precision and recall.

The classification performance on $F_1$ measure for the different classification datasets is shown in Figure \ref{fig:classfication}. 
First note that in the majority of the classification scenarios, \senLDA  outperforms LDA. In most cases, the performance difference increases when the larger train sets (``WikiTrain2'' and ``PubMedTrain2'') are used. For instance, in the second line of figures with the PubMed classification experiments, increasing the topic models' training data benefits both LDA and \senLDA, but \senLDA still performs better. 
More importantly though and in consistence with the perplexity experiments, the advantage of \senLDA remains: the faster \senLDA convergence benefits the classification performance. The \senLDA curves are steeper in the first training iterations and stabilize after roughly 30 iterations when the model converges. We believe that assigning the latent topics to coherent groups of words such as sentences results in document representations of finer level. In this sense,  spans larger than single words can capture and express the document's content more efficiently for discriminative tasks like classification. 

%

To investigate the correlation of topic model representations learned on different levels of text, we report the classification  performance using as document representations the concatenation of a document's topic distributions output by LDA and \senLDA. For instance, the concatenated vectorial representation of a document when $K=125$ for each model is a vector of 250 dimensions. The resulting concatenated representations are denoted by ``senLDA+'' in Figure \ref{fig:classfication}. As it can be seen, ``senLDA+'' performs better compared to both LDA and \senLDA. Its performance combines the advantages of both models: during the first iterations it is as steep as the \senLDA representations and in the later iterations benefits by the LDA convergence to outperform the simple \senLDA representation. Hence,  the concatenation of the two distributions creates a richer representation where the two models contribute complementary information that achieves the best classification performance. Achieving the optimal performance using those representations suggests that the relaxation of the independence assumptions between the text structural units can be beneficial; this is also among the contributions of this work.

\section{Conclusion}
We proposed \textit{senLDA}, an extension of LDA where topics are sampled per coherent text spans. This resulted in very fast convergence and good classification and perplexity performance. LDA and \senLDA differ in that the second assumes a very strong dependence of the latent topics between the words of sentences, whereas the first assumes independence between the words of documents in general. In our future research, our goal is to investigate this dependence and further adapt the sampling process of topic models to cope with the rich text structure.

\section{Acknowledgements} 
This work is partially supported by the CIFRE N 28/2015.

\bibliographystyle{abbrv}

\end{document}